%% file: bare_jrnl_new_sample_final_submission.tex
\documentclass[lettersize,journal]{IEEEtran}
\usepackage{amsmath,amsfonts}
\usepackage{algorithmic}
\usepackage{algorithm}
\usepackage{array}
\usepackage[caption=false,font=normalsize,labelfont=sf,textfont=sf]{subfig}
\usepackage{textcomp}
\usepackage{stfloats}
\usepackage{url}
\usepackage{verbatim}
\usepackage{graphicx}
\usepackage{cite}
\usepackage{multirow}
\usepackage{cleveref}
\begin{document}

\title{3D Dual-Fusion: Dual-Domain Dual-Query Camera-LiDAR Fusion for 3D Object Detection}

\author{Yecheol Kim$^1$, Konyul Park$^1$, Minwook Kim$^1$, Dongsuk Kum$^2$, and Jun Won Choi$^1$ \\
Hanyang University$^1$ \\ Korea Advanced Institute of Science \& Technology (KAIST)$^2$
\thanks{Yecheol Kim and Jun Won Choi are with Department of Electrical Engineering, Hanyang University, 04753 Seoul, Republic of Korea. (e-mail: yckim@spa.hanyang.ac.kr, junwchoi@hanyang.ac.kr) \textit{(Corresponding author: Jun Won Choi)}}
\thanks{Konyul Park and Minwook Kim are with Department of Artificial Intelligence, Hanyang University, 04753 Seoul, Republic of Korea. (e-mail: \{konyulpark, minwookkim\}@spa.hanyang.ac.kr)}
\thanks{Dongsuk Kum is with the Graduate School of Mobility, Korea Advanced Institute of Science and Technology(KAIST), 34141 Daejeon, Republic of Korea, (email: dskum@kaist.ac.kr)}}



\maketitle

\begin{abstract}
Fusing data from cameras and LiDAR sensors is an essential technique to achieve robust 3D object detection. One key challenge in camera-LiDAR fusion involves mitigating the large domain gap between the two sensors in terms of coordinates and data distribution when fusing their features. In this paper, we propose a novel camera-LiDAR fusion architecture called, 3D Dual-Fusion, which is designed to mitigate the gap between the feature representations of camera and LiDAR data. The proposed method fuses the features of the camera-view and 3D voxel-view domain and models their interactions through deformable attention. We redesign the transformer fusion encoder to aggregate the information from the two domains. Two major changes include 1) dual query-based deformable attention to fuse the dual-domain features interactively and 2) 3D local self-attention to encode the voxel-domain queries prior to dual-query decoding. The results of an experimental evaluation show that the proposed camera-LiDAR fusion architecture achieved competitive performance on the KITTI and nuScenes datasets, with state-of-the-art performances in some 3D object detection benchmark categories.
\end{abstract}

\begin{IEEEkeywords}
3D object detection, sensor fusion, transformer, autonomous driving, deep learning
\end{IEEEkeywords}

\begin{figure}[t]
	\centering
    \includegraphics[width=0.9\columnwidth]{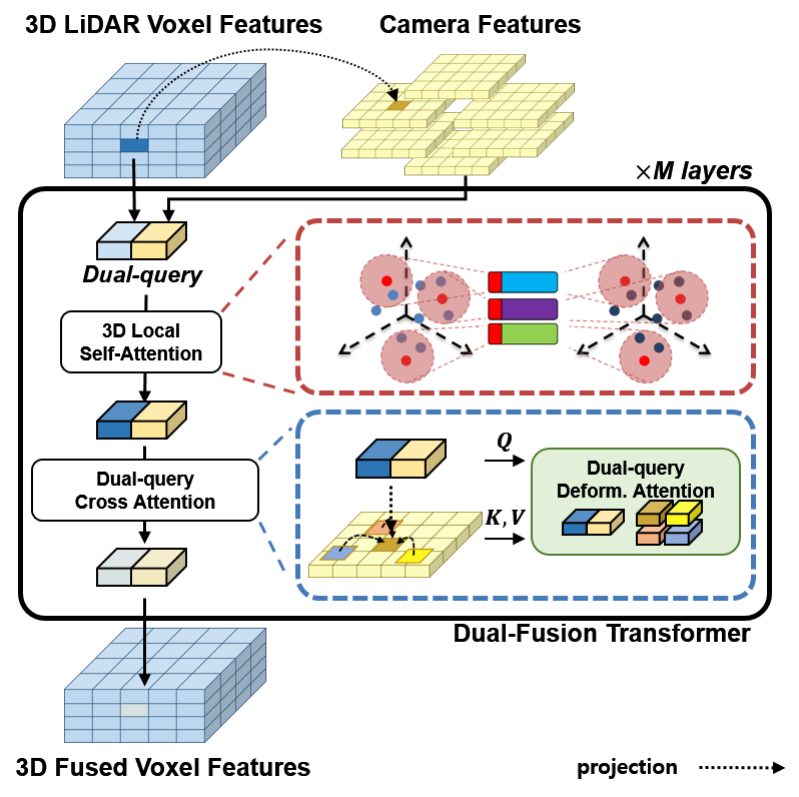}
	\caption {\textbf{Key concept of 3D Dual-Fusion:} 3D Dual-Fusion employs Dual-Fusion Transformer to enable dual-domain interactive feature fusion. Dual-Fusion Transformer first performs {\it 3D local self-attention} to encode the voxel-domain features. Then, it applies {\it dual-query cross attention}, which performs simultaneous feature fusion in both camera-view and voxel domains through dual-queries.     }
	\label{preoverall}
    \vspace{-0.5cm}
\end{figure}

\section{Introduction}
\IEEEPARstart{A} 3D object detector identifies the presence, location, and class of an object in a 3D world coordinate system based on sensor data. Both cameras and light detection and ranging (LiDAR) sensors provide useful information for 3D object detection. These two sensors have significantly different characteristics, because they use different physical sources and measurement processes. Camera sensors provide dense visual information on objects such as their color, texture, and shape, whereas LiDAR sensors produce accurate but relatively sparse range measurements. Consequently, these sensors exhibit different behavior and performance characteristics depending on the scene and object conditions. Camera-LiDAR sensor fusion aims to combine the complementary information provided by the two sensing modalities to achieve robust 3D object detection.

Recently, deep neural network (DNN) models have achieved considerable success in 3D object detection tasks, and numerous DNN architectures have been developed for LiDAR-based 3D object detection. These detectors encode LiDAR point clouds using   backbone networks and detect objects based on the features obtained from the encoder. {\it Voxel-based encoding} is a widely used LiDAR encoding method, that voxelizes LiDAR points in 3D space and encodes the points in each voxel \cite{voxelnet, second}. Camera images also provide useful information for 3D object detection. Visual features obtained by applying convolutional neural network (CNN) model to camera images can be used to perform 3D object detection as well \cite{bevformer}. To leverage diverse information provided by two multiple sensors, a {\it feature-level fusion strategy} has been proposed, which uses both voxel features and camera features together to perform 3D object detection.  However, these methods involve the challenge that two features are represented in different coordinate domains (i.e., camera-view versus voxel domains), so one of the coordinate representations must be aligned and adapted into another without losing information of the original domains. Therefore, the developement of method to reduce such a domain gap in aggregating  dual-domain features is a  key to improve the performance of camera-LiDAR sensor fusion methods.

Existing camera-LiDAR fusion methods used various domain transformation strategies at the point level, feature level, and proposal level. Point-level fusion methods \cite{mvxnet, vora2020pointpainting, ipod} projected the semantic information obtained from a camera image into the LiDAR points in 3D space and combined the data with point-wise LiDAR features. One limitation of these methods is that LiDAR data do not participate in fusion at the same semantic level as the camera features. Proposal-level fusion methods \cite{pointfusion, fpointnet, ipod} generated 2D detection proposals from a camera image and associated LiDAR points with each proposal. Then, the camera features and LiDAR features were fused to refine each proposal. However, the performance of these methods was limited by the accuracy of the generated proposals. Feature-level fusion methods \cite{mmf, contfuse, pointaugmenting, epnet, 3d-cvf,transfusion, autoalign} are designed to extract semantic features separately from the camera image and LiDAR data and aggregate them in the voxel domain. MMF \cite{mmf}, ContFuse \cite{contfuse}, PointAugmenting \cite{pointaugmenting}, EPNet \cite{epnet}, and 3D-CVF \cite{3d-cvf} transformed the image features into a voxel domain using a calibration matrix and conducted the element-wise feature fusion.  Transfusion  \cite{transfusion} and AutoAlign \cite{autoalign} employed the attention mechanism of Transformer to incorporate a relevant part of the camera features in the camera-view domain to the voxel domain.

In this paper, we aim to reduce the domain gap between the camera features and LiDAR features to boost the effect of sensor fusion. We  propose a new camera-LiDAR fusion architecture, referred to as {\it 3D Dual-Fusion}, for 3D object detection.  The key idea of the proposed approach is {\it dual-domain interactive feature fusion}, in which the camera-domain features and voxel-domain LiDAR features are transformed into each other's domains and simultaneously fused in each domain. To this end, we design a Dual-Fusion Transformer architecture consisting of {\it 3D local self-attention} and {\it dual-query deformable attention}. First, 3D local self-attention applies self-attention to encode the voxel-domain features in each local group of non-empty voxels.  {\it Dual-query deformable attention} uses two types of queries (called {\it dual-queries}) to progressively refine both the camera-domain features and voxel-domain features through multiple layers of deformable attention.  

This strategy can preserve information that might have been lost if the feature transformation is performed once in a particular stage and subsequent interactions between additional domains are not performed. Such collaborative feature fusion  between the camera-view and voxel domains produces strong dense bird's eye view (BEV) feature maps, which can be processed by any {\it off-the-shelf} 3D object detection head.  The key concept of the proposed dual-domain feature fusion is illustrated in Fig. \ref{dda}. 

We evaluate the performance of 3D Dual-Fusion on two widely used public datasets including KITTI dataset \cite{kitti} and nuScenes dataset \cite{nuscenes}. We show that the proposed 3D Dual-Fusion achieves a significant performance improvement over other camera-LiDAR fusion methods and records competitive performance on official KITTI and nuScenes leaderboards.
 
 The contributions of this study are summarized as follows:
\begin{itemize}
    \item We propose a novel dual-query deformable attention architecture for camera-LiDAR sensor fusion. The original deformable attention structure is limited in supporting cross attention for fusing multi-modal features. To address this, we introduce the concept of a dual-query, which can enable simultaneous cross-domain feature aggregation through deformable attention.

    \item We leverage the sparsity of LiDAR points to reduce the computational complexity and memory usage of 3D Dual-Fusion. We assign dual-queries only for the non-empty voxels and the corresponding pixels of the camera features projected from those voxels. Because only a small fraction of voxels are non-empty, the number of queries used for dual-domain interactive feature fusion is much smaller than the size of the entire voxel.  Furthermore, adopting the structure of deformable attention, the 3D Dual-Fusion architecture is lightweight compared to global-scale attention in the Transformer architecture. 
     \item When combined with the TransFusion detection head \cite{transfusion}, the proposed 3D Dual-Fusion method achieves  {\it state-of-the-art} performance  on some categories of KITTI and nuScenes benchmarks.
    
 \item The source codes used in this work will be released publicly. 
\end{itemize}

\begin{figure*}[t]
	\centering
        \centerline{\includegraphics[width=0.9\textwidth]{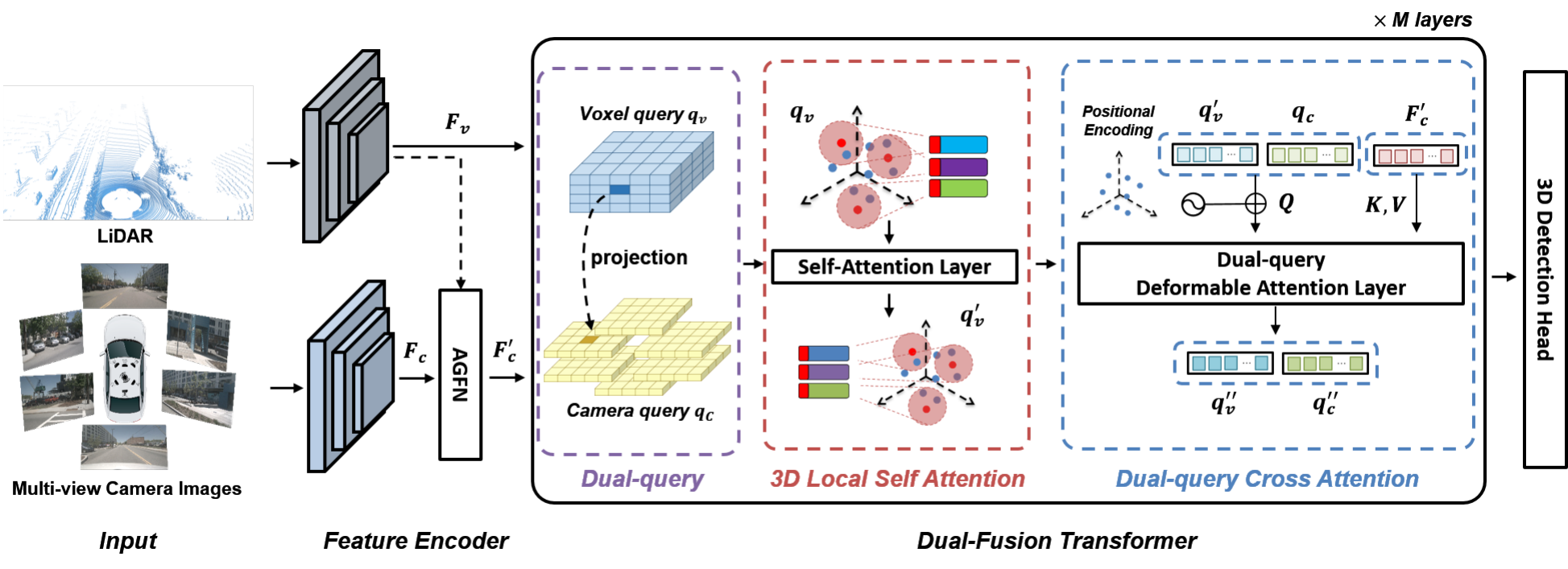}}
    	\caption {\textbf{Overall structure of 3D Dual-Fusion.} Camera-domain features and voxel-domain features are obtained through their own backbone networks. First, camera-domain features are enhanced by the camera-domain feature fusion through AGFN. Then, Dual-Fusion Transformer is applied to perform simultaneous dual-domain feature fusion using dual-queries. Finally, the final voxel-domain features are returned to the 3D detection head to generate the detection results.}
	\label{overall}
\end{figure*}

\section{Related Work}
\label{sec:Related work}

\subsection{LiDAR-only 3D Object Detection}

Numerous architectures have been proposed to perform 3D object detection based on LiDAR point clouds. Two well-known backbone networks have been developed to encode LiDAR data, including voxel-based \cite{voxelnet,second,PointPillars} and PointNet++-based backbones \cite{pointnet++,pointrcnn,3dssd}. The voxel-based backbone partitions the LiDAR points using a voxel or a pillar structure and encodes the points of each grid element. In contrast, the PointNet++-based backbone groups the points using the farthest point sampling algorithm and increases the semantic level of the point-wise features in a hierarchical manner. 
Recently, Transformer models have been used to encode LiDAR point clouds \cite{ct3d, votr}.

\subsection{Camera-LiDAR Fusion for 3D Object Detection}
To date, various camera-LiDAR sensor fusion methods have been proposed to achieve robust 3D object detection. 
These methods can be roughly categorized into point-level, feature-level, and proposal-level fusion methods. 
 
Point-level fusion augments the semantic information obtained from a camera image to point-wise features extracted from LiDAR data. MVXNet \cite{mvxnet} transformed the visual features extracted from a 2D object detector, while PointPainting \cite{vora2020pointpainting} and FusionPainting \cite{fusionpainting} transferred the semantic segmentation masks to the LiDAR points. 


Proposal-level fusion relies on the detection proposals obtained by processing a single sensor to achieve sensor fusion.  RoarNet \cite{roarnet}, F-PointNet, \cite{fpointnet} and PointFusion \cite{pointfusion} predicted 2D detection proposals based on a camera image and associated the corresponding LiDAR points using proposal frustums. MV3D \cite{mv3d} and AVOD \cite{avod} predicted 3D detection proposals using LiDAR data and gathered the camera and LiDAR features corresponding to each 3D box proposal.  


Feature-level fusion extracts features from data collected by two sensors and aggregates them through a domain transformation. EPNet \cite{epnet} and 3D-CVF \cite{3d-cvf} enhanced the semantic information of LiDAR features by transforming the image features obtained at different levels of scale to the camera domain. MMF \cite{mmf} and ContFuse \cite{contfuse} used continuous convolution to fuse multi-scale convolutional feature maps obtained from each sensor. VFF \cite{vff} proposed a ray-wise fusion to utilize the supplemental context in the voxel field. Recently, DeepFusion \cite{deepfusion}, AutoAlign \cite{autoalign}, and TransFusion \cite{transfusion} used the Transformer architecture to dynamically capture the correlations between camera and LiDAR features.

\subsection{Review of Deformable DETR}

Deformable DETR \cite{def-detr} realized Transformer attention \cite{transformer} at low computational complexity using a local attention mechanism. 
Deformable attention only attends to a small set of key sampling points around a reference point. For the sake of brevity, we skip the multi-scale term of deformable kernels. Given an input feature map $F_{c}\in\mathbb{R}^{H{\times}W{\times}C}$, let $q$ index a query with feature $z_{q}$ and the reference point $p_{q}$. The deformable attention is formulated by
\begin{align}
   & DefAttn\left(z_q,p_q,F_{c}\right)= \nonumber \\ 
   & \;\;\; \sum_{m=1}^{M}{W_m\sum_{k=1}^{K}{A_{mqk}\cdot W_m^\prime F_{c}\left(p_q+\mathrm{\Delta}p_{mqk}\right)}},
\end{align}
where $m$ indexes the attention head, $k$ indexes the sampled keys, and $K$ is the total number of the sampled keys. $W_{m}\in\mathbb{R}^{C{\times}C/M}$ and $W^{\prime}_m\in\mathbb{R}^{C/M{\times}C}$ denote learnable projection matrices. The attention weight $A_{mqk}$ predicted from the query feature $z_q$ is in the range $[0,1]$, and $\sum_{k=1}^{K} A_{mqk}=1$. $\Delta p_{mqk}$ is the predicted offset from the reference point, and $F_{c}(p_{q}+\Delta p_{mqk})$ is the input feature at the $p_q+\Delta p_{mqk}$ location. In this study, we tailor deformable DETR to support the proposed dual-domain fusion of camera and LiDAR features.

\begin{figure*}[tbh]
	\centering
        \centerline{\includegraphics[width=0.65\textwidth]{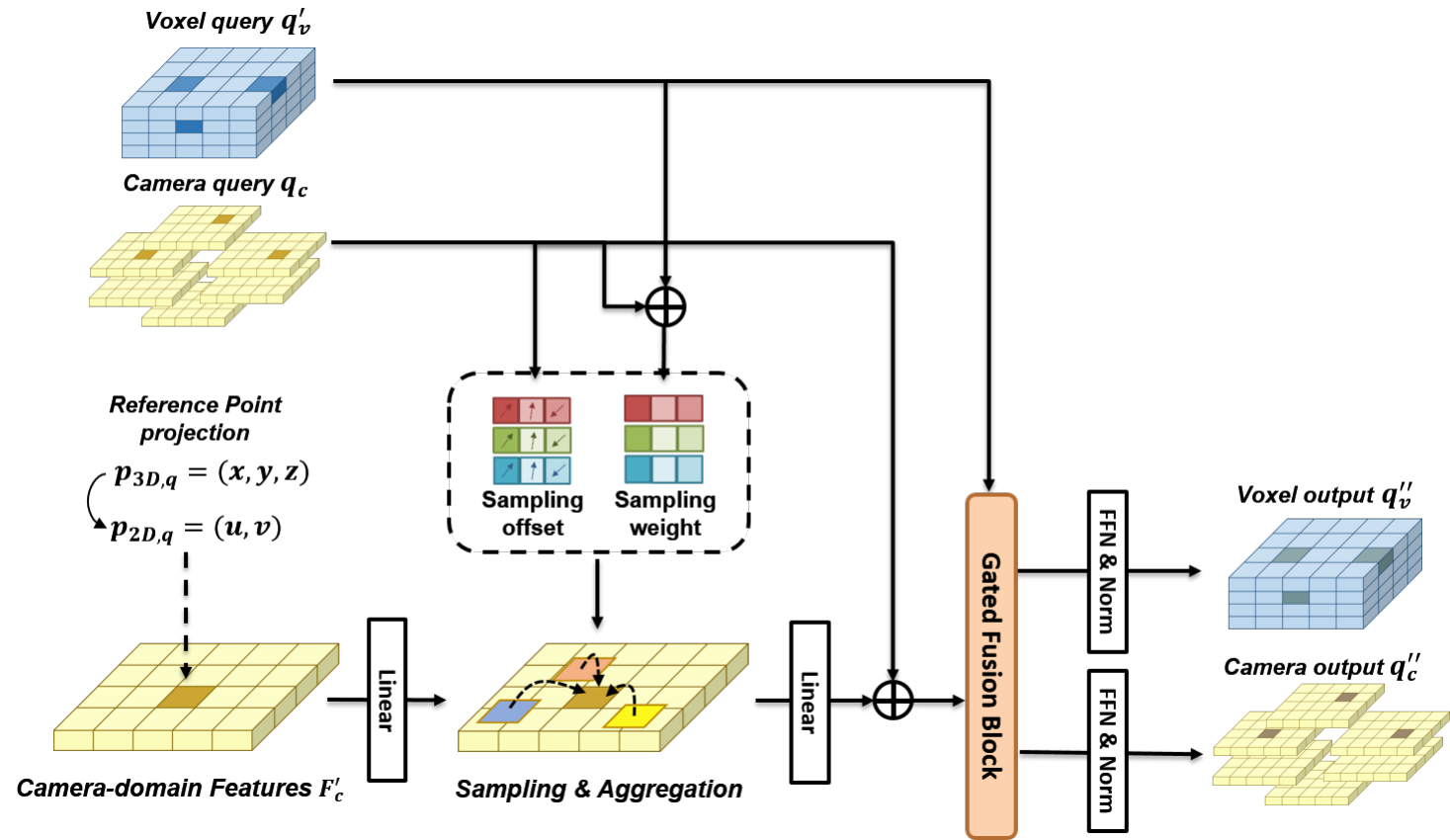}}
    	\caption {\textbf{Structure of Dual-query Deformable Attention  (DDA):} DDA decodes dual-queries to perform dual-domain interactive feature fusion. First, DDA updates the c-queries $q_c$ by $q'_c$ applying the deformable attention on the camera-domain features $F'_c$. Then, the gated fusion block fuses $q'_c$ and  $q'_v$ with adaptive ratios to update the queries to $q''_c$ and $q''_v$. }
	\label{dda}
\end{figure*}

\section{3D Dual-Fusion}

In this section, we present the details of 3D Dual-Fusion. 

\subsection{Overall Architecture}
The overall architecture of the proposed 3D Dual-Fusion is illustrated in Fig \ref{overall}. The camera image $X_{c}\in\mathbb{R}^{{W_{c}}{\times}{H_{c}}{\times}3}$ is encoded by the standard CNN backbone, where $W_{c}$ and $H_{c}$ denote the width and height of the camera image, respectively. In multi-view camera setup, multiple camera images are encoded by CNN separately. LiDAR point clouds $X_{v}\in\mathbb{R}^{N{\times}3}$ are also encoded by voxel encoding backbone, where $N$ is the number of LiDAR points. We voxelize the LiDAR points using the voxel grid structure of the width $W_{v}$, length $L_{v}$ and height $H_{v}$.  The structure of the voxel encoding backbone is adopted from  \cite{second}. We also use the CNN backbone of  DeepLabV3 \cite{deeplabv3}. Voxel encoding backbone network produces the voxel-domain LiDAR features $F_{v}\in\mathbb{R}^{{W_{v}}{\times}{L_{v}}{\times}{H_{v}}{\times}{C_{v}}}$. The CNN backbone network produces the camera-domain image features $F_c\in\mathbb{R}^{{W_{c}}/8{\times}{H_{c}}/8{\times}{C_{c}}}$.


The voxel-domain LiDAR features $F_{v}$ and camera-domain image features $F_{c}$ are fed into the Dual-Fusion Transformer. Before that, the camera-domain image features $F_{c}$ are enhanced by an {\it adaptive gated fusion network} (AGFN). AGFN projects the voxel-domain LiDAR features to the camera domain and combines them with the camera-domain features. We denote the features enhanced by AGFN as $F'_{c}$.
Dual-Fusion Transformer applies {\it 3D local self-attention} (3D-LSA) and {\it dual-query deformable attention} (DDA) to fuse  $F_{v}$ and $F'_{c}$. The voxel-domain features in non-empty voxels are first encoded by {\it 3D local self-attention}. Then, DDA performs simultaneous feature fusion in both camera-view and voxel domains through cross attention with dual-queries.  
After multiple attention layers, the final voxel-domain features are transformed into bird's eye view (BEV) features \cite{second}.  BEV features are passed through the detection head to generate a 3D bounding box and classification score.
Note that the entire network of 3D Dual-Fusion is end-to-end trainable.

\subsection{Dual-Query}
The dual queries consist of a camera-query ({\it c-query}) and  voxel-query ({\it v-query}).
Suppose that the LiDAR voxelization step yields $Q$ non-empty voxels, where $Q$ can vary depending on the distribution of the input point clouds.  v-queries $q_v = \{q_{v,1},q_{v,2},..., q_{v,Q}\}$ are assigned to the non-empty voxels and used to refine their voxel-domain features.  c-queries $q_c = \{q_{c,1},q_{c,2},..., q_{c,Q}\}$ are assigned to  the image pixels indicated by projecting the center points of the non-empty voxels into the camera domain and used to refine the image-domain features.  


\subsection{3D Local Self-Attention}
In the beginning, v-queries are initialized with the LiDAR features in non-empty voxels. The self-attention layer successively encodes v-queries by modeling their spatial relationships. Because global self-attention is computationally intensive,  {\it 3D local self-attention} is devised to reduce the scope of attention within a local region. The non-empty voxels are clustered into local regions by applying the {\it farthest point sampling} algorithm \cite{pointnet++} to their center points. We find $K$ voxels within a fixed radius around a centroid of each local region. Let $F=\{f_i|i\in{N(x_{ct})}\}$ and $X=\{x_i|i\in{N(x_{ct})}\}$ be a set of query features and 3D positions assigned to the centroid $x_{ct}$, respectively. Then, 3D-LSA module performs self-attention as
\begin{align}
PE(x_i, x_j) &= FFN(x_i - x_j)  \nonumber \\
q_i^{(l)}&={f_i^{(l)}}{W_q}, \quad k_i^{(l)} = {f_i^{(l)}}{W_k}, \quad, v_i^{(l)}={f_i^{(l)}}{W_v} \nonumber \\
y_i^{(l)} &= \sum_{j\in N(x_{ct})}{\rm softmax}({q_i}{k_j} / \sqrt{d} + PE(x_i, x_j)) \nonumber  \\
f_{i}^{(l+1)} &= f_i^{(l)} + FFN(y_i^{(l)}), 
\end{align}
where $W_q$, $W_k$, and $W_v$ are the projection matrices for query, key and value, respectively, $l$ is the index of $L$-layer Transformer block and $d$ is the scaling factor for normalizing dot-product attention. $FFN(\cdot)$ denotes a position-wise feed forward network, and $PE(x_i, x_j)$ is the positional encoding function that encodes the difference of two 3D coordinates  $x_i$ and $x_j$ through FFN. 
 If the number of non-empty voxels in the radius is larger than $K$, we randomly choose $K$ voxels. The v-queries associated with a group of $K$ voxels are separately encoded by self-attention \cite{transformer}. 
After multiple self-attention layers, the input v-queries $q_{v} = \{q_{v,1},q_{v,2},..., q_{v,Q}\}$ are updated by the new v-queries $q'_{v} =  \{q'_{v,1},q'_{v,2},..., q'_{v,Q}\}$.


\subsection{Dual-Query Cross Attention}
The proposed DDA is designed to perform dual-domain feature fusion efficiently. The original transformable attention \cite{def-detr} supports only attention on a single modality, so some modifications are needed to support joint attention on both 2D and 3D modalities. 

The structure of DDA is shown in Fig. \ref{dda}. Let $q$ index $Q$ non-emtpy voxels. Consider a 3D reference point $p_{3D,q}=(x,y,z)$ at the center point of the $q$th non-empty voxel. The corresponding 2D reference point $p_{2D,q}=(u,v)$  is determined by projecting $p_{3D,q}$ on the camera domain and quantizing it on the pixel grid. In multi-view camera setup, $p_{3D,q}$ can be projected on multiple images. In this case, only a single image is chosen for projection. For consistency, we choose the camera to the far right of the ego vehicle's direction of travel.
The c-query $q_{c,q}$ is initialized by the camera-domain features $F'_c$ indicated by the reference point $p_{2D,q}$. The v-query $q^{\prime}_{v,q}$ is obtained from 3D local self-attention. 

The depth-aware positional encoding is first applied to both v-queries and c-queries. Rather than using $(u,v)$-based encoding in the original deformable DETR, the positional embedding is computed based on the depth $x$ of $p_{3D,q}$ 
\begin{align}
PE_{(q,2i)}=\sin(x/(10000^{2i/d_{model}})) \\
PE_{(q,2i+1)}=\cos(x/(10000^{2i/d_{model}})),
\end{align}
where  $i$ and $d_{model}$ are the index and dimension of the query vector, respectively. This depth-aware positional embedding is then added to both queries.

 \input{table/kitti_test.tex}

The dual-query cross attention decodes the dual-queries $q^{\prime}_{v}$ and $q_{c}$ over multiple attention layers.
First, deformable attention is applied to transform the c-queries $q_{c}$ using the camera features $F'_{c}$  as key and value. For a given 2D reference point $p_{2D,q}=(u,v)$, a deformable mask with adaptive offsets and weights is applied to the camera-domain features $F'_c$. The mask offsets $ \Delta p_{mqk}$ and  mask weights $A_{mqk}$ are determined as
\begin{align}
 \Delta p_{mqk}=FFN\left(q_{c,q}\right),\ A_{mqk}=FFN\left(q_{c,q}+q^\prime_{v,q}\right),
\end{align}
where  $+$ denotes the element-wise summation and FFN denotes the feed forward network. Note that the mask weights are determined using both v-queries and c-queries. This design is intended to boost the effect of feature fusion by determining the attention weights based on voxel-area and camera-area features.  
Given the offset $\Delta{p}_{mqk}$ and weight ${A}_{mqk}$, the attention value $q^\prime_c = [q^\prime_{c,1},...,q^\prime_{c,Q}]$ is computed as 
\begin{align}
    q^\prime_{c,q}=   \sum_{m=1}^{M}{W_m\sum_{k=1}^{K}{A_{mqk}\cdot W_m^\prime F^\prime_c\left(p_{2D,q}+\mathrm{\Delta}p_{mqk}\right)}},
\end{align}
where $m$ indexes the attention head, $k$ indexes the sampled keys, and $K$ is the total number of the sampled keys. $W_{m}\in\mathbb{R}^{C{\times}C/M}$ and $W^{\prime}_m\in\mathbb{R}^{C/M{\times}C}$ denote learnable projection matrices.

The queries $q^\prime_v$ and $q^\prime_c$ are further transformed by the gated fusion mechanism \cite{gatedfusion}. That is, $q^\prime_c$ and $q^\prime_v$ are fused with different ratios given by 
\begin{gather}
q^{\prime\prime}_c=q^\prime_c+q^\prime_v\times\sigma({Conv}_1(q^\prime_c+q^\prime_v)) \\
q^{\prime\prime}_v=q^\prime_v+q^\prime_c\times\sigma({Conv}_2(q^\prime_c+q^\prime_v)),
\end{gather}
where $\times$ denotes the element-wise multiplication,  $\sigma(\cdot)$ is the sigmoid function $\frac{1}{1+e^{-x}}$, and ${Conv}_1(\cdot)$ and ${Conv}_2(\cdot)$ are the convolutional layers with different weights. Note that the combining ratios adjust adaptively to the input features since they are a function of $q^\prime_v$ and $q^\prime_c$.  Finally, DDA produces the decoded queries $q^{\prime\prime}_c$ and $q^{\prime\prime}_v$. They are used as input queries for the next Dual-Fusion Transformer layer. 



\subsection{Adaptive Gated Fusion Network}
 AGFN projects LiDAR features to the camera domain and fuses the projected features with camera features. AGFN also combines two features using the gated fusion mechanism \cite{gatedfusion}.  The AGFN produces the features $F^\prime_c$ through the following operations
\begin{gather}
A=T(F_v)\times\sigma({\rm Conv}_v(T(F_v)+F_c))\\
B=F_c\times\sigma({\rm Conv}_c(T(F_v)+F_c))\\
F^\prime_c={Conv}_r(A \oplus B ),
\end{gather}
where $T( \cdot)$ denotes the projection on the camera domain and $\oplus$ denotes the concatenation operation.

\section{Experiments} 
\label{Experiments}
In this section, we evaluate the performance of 3D Dual-Fusion on  KITTI dataset \cite{kitti} and  nuScenes dataset \cite{nuscenes}. 
\subsection{Experimental setup}

\subsubsection{KITTI Dataset}
KITTI dataset is a widely used dataset for 3D object detection task. The dataset was collected with a vehicle equipped with a 64-beam Velodyne LiDAR point cloud and a single PointGrey camera. The dataset comprises 7,481 training samples and 7,518 testing samples. The training samples are commonly split into a training set with 3,712 samples and a validation set with 3,769 samples. In the evaluation, we used the {\it mean average precision} (mAP) as a primary metric. mAP was calculated using recall 40-point precision (R40) and 11-point precision (R11).   We used mAP metric $AP_{3D}$ for 3D object detection task and mAP metric $AP_{BEV}$ for BEV object detection task. 

\subsubsection{nuScenes Dataset}
nuScenes dataset is a large-scale benchmark dataset for autonomous driving. The dataset provides the samples from a synced 32-beam LiDAR, six cameras, and five radars covering 360 degrees. It contains 1,000 scenes collected from Boston and Singapore, which consists of 700 training scenes, 150 validation scenes, and 150 testing scenes.  Each scene was recorded over 20 seconds at 2Hz and  10 classes of objects were annotated.  Models were evaluated mainly using the {\it nuScenes detection score} (NDS) \cite{nuscenes} and mAP. For the ablation study, 3D Dual-Fusion was trained on  1/7 of the training data and evaluated on the full validation dataset.

\input{table/nus_test.tex}

\input{table/abl_overall.tex}

\subsubsection{Implementation Details}
For the KITTI dataset, we used the backbone network and detection head of  Voxel R-CNN \cite{voxelrcnn} for 3D Dual-Fusion. The range of the point cloud data was within $[0, 70.4]\times[-40, 40]\times[-3, 1]m$ on  $(x, y, z)$ axis. The point clouds were voxelized with each voxel size of $0.05\times0.05\times0.1m$. The number of Dual-Fusion layers, $M$ was set to 4. We used DeepLabV3 \cite{deeplabv3} pre-trained on the COCO data for the camera backbone.
We used the data augmentation in Voxel R-CNN \cite{voxelrcnn}. In addition, we used  the point-image GT sampling augmentation \cite{moca}. 

For the nuScenes dataset, we used two baseline networks: CenterPoint \cite{centerpoint} and  TransFusion \cite{transfusion}. Note that CenterPoint is LiDAR-only based detector and TransFusion is camera-LiDAR fusion-based detector.  We integrated the backbone network and detection head of these methods to 3D Dual-Fusion. The detection head of TransFusion performs proposal-level feature fusion. We applied the detection head of TransFusion to the dense feature maps produced by 3D Dual-Fusion.  3D Dual-Fusion with CenterPoint baseline is called {\it 3D Dual-Fusion (C)} and 3D Dual-Fusion with TransFusion baseline is called {\it 3D Dual-Fusion (T)}. The range of the point clouds was within $[-54, 54]\times[-54, 54]\times[-5, 3]m$ on  $(x,y,z)$ axis. The point clouds were voxelized with a voxel sizes of $0.075\times0.075\times0.2m$. We used six camera images with multiple views to extract camera features. The number of Dual-Fusion layers $M$ was set to 2. Point-image GT sampling was not enabled in the last four epochs. 


\subsection{Main Results}
\subsubsection{KITTI} We evaluate the performance of the proposed 3D Dual-Fusion  on {\it KITTI test set}. Table \ref{table:kitti_test} compares the detection accuracy of 3D Dual-Fusion with that of other top ranked 3D object detectors in the leaderboard. We included both  LiDAR-only and camera-LiDAR fusion-based methods in our comparison. We did not include the methods using multi-frame sequence data.  The performance of other 3D detection methods was obtained from the official KITTI leaderboard. Table \ref{table:kitti_test} shows that for 3D object detection task, 3D Dual-Fusion achieves the best performance among candidates in all the easy, moderate, and hard categories. For the BEV object detection task, 3D Dual-Fusion achieves the best performance only in the moderate category and competitive performance in other categories. In particular, 3D Dual-Fusion sets a new state-of-the-art performance surpassing the current best method, Focals Conv. Compared to  the LiDAR-only based baseline, Voxel R-CNN, the proposed sensor fusion method yields improvements of 0.11\%, 0.78\%, and 2.33\% on the easy, moderate, and hard categories of the 3D object detection task, respectively. Notably, the performance gain is much larger for the hard category, which shows that sensor fusion is particularly effective in detecting difficult objects. 


\subsubsection{nuScenes} We also evaluated the performance of 3D Dual-Fusion on {\it nuScenes test set} and {\it val set}. Table \ref{table:nus_test} shows a comparison of the detection accuracy of top 3D object detectors in the nuScenes leaderboard. On top of mAP and NDS, we use AP metrics for each object class category. Both 3D Dual-Fusion (T) and 3D Dual-Fusion (C) are competitive with the latest 3D object detection methods. Particularly, 3D Dual-Fusion (T) outperforms all other 3D object detection methods in both mAP and NDS metrics. 3D Dual-Fusion (T) achieves 0.4\%  performance gain in mAP and 0.2\% gain in NDS over the current state-of-the-art BEVFusion.  3D Dual-Fusion (C) offers 4.2\% gain in NDS and 7.6\% gain in  mAP over CenterPoint baseline \cite{centerpoint}.  3D Dual-Fusion (T) offers  1.7\% gain in mAP and 1.4\% gain in NDS over Transfusion baseline \cite{transfusion}. This performance improvement is achieved by providing dense feature maps enhanced by the proposed dual-domain fusion to the detection head of TransFusion.




\input{table/nus_val.tex}

\input{table/abl_nus_dist.tex}

\input{table/abl_position_encoding.tex}
\input{table/abl_num_encoder_layer_kitti.tex}
\input{table/abl_num_encoder_layer_nus.tex}
\subsection{Performance Analysis}
\subsubsection{Ablation Study}
 Table \ref{table:abl_overall} evaluates the contributions of each module of 3D Dual-Fusion to its overall performance.   We provide the evaluation results on both KITTI and nuScenes datasets. CenterPoint \cite{centerpoint} is the baseline for the nuScenes dataset and Voxel R-CNN \cite{voxelrcnn} is the baseline for the KITTI dataset. 
 We consider the following components: (a) \textit{naive fusion}, (b) DDA, (c) 3D local self-attention (3D-LSA), and (d) AGFN. 
 First, we consider a {\it naive fusion} method that brings the camera features associated with the non-empty voxels  and fuses them with LiDAR voxel features through summation. We see the \textit{naive fusion} improves mAP in the moderate category by 0.62\% on KITTI and NDS by 1.9\% on nuScenes. 
 Then, we incorporate the proposed fusion method, DDA.  
 DDA yields  0.47\% additional mAP gain in the moderate category on KITTI and 2.0\% NDS gain on nuScenes. DDA achieves a considerable performance improvement over the baseline, which  
demonstrate the effectiveness of the proposed dual-domain fusion method.  Next, we add 3D-LSA on top of DDA. 3D-LSA offers 0.29\% mAP gain in the moderate category on KITTI and 0.7\% NDA gain on nuScenes. The performance gains due to 3D-LSA are relatively small, but they contribute to surpassing other methods. Finally, the addition of AGFN provides an additional gain of 0.3\% NDA on nuScenes, which is not negligible. AGFN achieved a 0.34\% improvement in the easy category but relatively little on other categories on KITTI. 

Table \ref{table:abl_nus_dist} shows the performance of 3D Dual-Fusion (C) evaluated for objects present in three different distance ranges. We compare the performance of 3D Dual-Fusion with that of LiDAR-based CenterPoint baseline \cite{centerpoint}. We categorized the objects in nuScenes val split into three distance categories $(<15m)$, $(15\sim30m)$, and $(>30m)$. We observe that 3D Dual-Fusion achieves 5.5\% gain in NDS for $(15\sim30m)$ and $(>30m)$ cases, whereas it achieves 2.1\% gain for $(<15m)$ case. 
 This result shows that sensor fusion is more effective for more distant objects.

\subsubsection{Positional Encoding}
Table \ref{table:abl_pe} shows the effectiveness of the depth-aware positional encoding method. We compared our depth-based positional encoding with the $(u,v)$-based positional encoding used in the original deformable DETR. Table \ref{table:abl_pe} shows that the proposed depth-aware positional encoding achieves a 1.5\% improvement in mAP and 1.6\% gain in NDS over the $(u,v)$-based encoding. This appears to be because the proposed positional encoding provides a sense of depth to DDA, allowing to  utilize visual context features better.

\input{table/abl_query_type_nus.tex}
\input{table/abl_dda_fusion_method.tex}
\subsubsection{Analysis of DDA Behavior}
Table \ref{table:abl_num_encoder_layer_kitti} and Table \ref{table:abl_num_encoder_layer_nus} show the performance trend with respect to the number of decoder layers on KITTI {\it validation} set and nuScenes {\it validation} set, respectively. In the KITTI dataset, performance gradually improves as the number of layers increases to 4. We could not try more than 4 layers due to limitation of GPU memory. In the nuScenes dataset, the performance rapidly improves in the first two layers and stops afterwards. This shows that our dual-query attention progressively refines the quality of the features used for 3D object detection.  

Table \ref{table:abl_query_type} compares our dual-query with the {\it single camera query}. The {\it single camera query} implies that only camera queries are updated through deformable attention. Specifically, the deformable mask offset and weights are determined by the camera queries only. Without the gated fusion, DDA produces the updated camera queries by adding the output of deformabe attention to the voxel queries. Note that the dual-query outperforms the {\it single camera query} by 2.4\% in mAP and 0.9\% in NDS on nuScenes.

In Table \ref{table:abl_dda_fm}, we investigate the benefit of using the gated fusion block in DDA. We compare different query fusion methods: 1) concatenation, 2) summation, and 3) gated fusion. We observe that the gated fusion block achieves a significant performance gain over other two query fusion methods.


\section{Conclusions}
In this paper, we proposed a novel camera-LiDAR fusion method for 3D object detection, called 3D Dual-Fusion. We introduced the concept of {\it dual-domain feature fusion}, in which the camera and LiDAR features  are aggregated in both voxel and camera domains in an interactive manner. To this end, we designed a 3D Dual-Fusion Transformer, which employed dual queries to aggregate the relevant  camera and LiDAR features through multiple attention layers. We designed a novel {\it dual-query cross attention} to enable simultaneous dual-domain feature fusion by decoding dual-queries. We also added {\it 3D local self-attention} and {\it adaptive gated fusion network} for further performance enhancement. Our evaluation conducted on KITTI and nuScenes datasets confirmed that the proposed ideas offered a significant performance boost over the baseline methods and that the proposed 3D Dual-Fusion achieved the state-of-the-art performance in the benchmarks.

{
\bibliographystyle{IEEEtran}
\bibliography{egbib}
}

\vfill

\end{document}

%% file: table/kitti_test.tex
\begin{table*}[t]
\setlength{\tabcolsep}{11pt}
\renewcommand{\arraystretch}{1.0}
\centering
\caption{Performance comparisons of 3D object detectors on KITTI \textit{test} set.}
\begin{tabular}{ccccccc}

\cline{1-7}
\multirow{2}{*}{Method} &  \multicolumn{3}{c}{$AP_{3D}$@Car-R40 (IoU=0.7)} & \multicolumn{3}{c}{$AP_{BEV}$@Car-R40 (IoU=0.7)} \\
\cline{2-4}
\cline{5-7}
                          & Easy     & \textbf{Moderate}    & Hard     & Easy      & \textbf{Moderate}    & Hard     \\
\cline{1-7}
\multicolumn{7}{c}{\textit{LiDAR based}}                                                                                \\
\cline{1-7}
PV-RCNN\cite{pvrcnn}           & 90.25    & 81.43                & 76.82    & 94.98     & 90.65                & 86.14    \\
Voxel R-CNN\cite{voxelrcnn}       & 90.90    & 81.62                & 77.06    & 94.85     & 88.63                & 86.13    \\
VoTr-TSD\cite{votr}      & 89.90    & 82.09                & 79.14    & 94.03     & 90.34                & 86.14    \\
Pyramid-PV\cite{pyramid}        & 88.39    & 82.08                & 77.49    & 92.19     & 88.84                & 86.21    \\
IA-SSD\cite{iassd}            & 88.87    & 80.32                & 75.10    & 93.14     & 89.48                & 84.42    \\
PDV\cite{pdv}           & 90.43    & 81.86                & 77.36    & 94.56     & 90.48                & 86.23    \\
\cline{1-7}
\multicolumn{7}{c}{\textit{LiDAR-Camera based}}                                                                                  \\
\cline{1-7}
MMF\cite{mmf}               & 86.81    & 76.75                & 68.41    & 89.49     & 87.47                & 79.10    \\
EPNet\cite{epnet}         & 89.81    & 79.28                & 74.59    & \textbf{94.22}     & 88.47                & 83.69    \\
3D-CVF\cite{3d-cvf}            & 89.20    & 80.05                & 73.11    & 93.52     & 89.56                & 82.45    \\
CLOCs\cite{clocs}             & 88.94    & 80.67                & 77.15    & 93.05     & 89.80                & \textbf{86.57}    \\
HMFI\cite{hmfi} & 88.90 & 81.93 & 77.30 & 93.04 & 89.17 & 86.37 \\
VFF\cite{vff}           & 89.50    & 82.09                & 79.29    &   -       &   -                  &   -      \\
Focals Conv\cite{focalsconv}       & 90.55    & 82.28                & 77.59    & 92.67     & 89.00                & 86.33    \\
\cline{1-7}
\textbf{3D Dual-Fusion} & \textbf{91.01}    & \textbf{82.40}       & \textbf{79.39}& 93.08& \textbf{90.86}       & 86.44\\
\cline{1-7}
\label{table:kitti_test}
\end{tabular}
\end{table*}

%% file: table/nus_test.tex
\begin{table*}[t]
\renewcommand{\arraystretch}{1.0}
\centering
\caption{Performance comparisons of 3D object detectors on nuScenes \textit{test} set. }
\begin{tabular}{cccccccccc}
\cline{1-10}
\textit{Method}        & mAP  & NDS     & Car    & Truck   & C.V.  & Ped.  & Motor & Bicycle  & Barrier \\
\cline{1-10}
\multicolumn{10}{c}{\textit{LiDAR-based}}                                                                \\
\cline{1-10}
PointPillars\cite{PointPillars}  & 40.1 & 55.0  & 76.0  & 31.0   & 11.3  & 64.0 & 34.2 & 14.0    & 56.4   \\
CenterPoint\cite{centerpoint}   & 60.3 & 67.3  & 85.2  & 53.5   & 20.0  & 84.6 & 59.5 & 30.7    & 71.1   \\
TransFusion\cite{transfusion}   & 65.5 & 70.2  & 86.2  & 56.7   & 28.2  & 86.1 & 68.3 & 44.2    & 78.2   \\
\cline{1-10}
\multicolumn{10}{c}{\textit{LiDAR-Camera based}}                                                               
\\
\cline{1-10}
3D-CVF\cite{3d-cvf}        & 52.7 & 62.3  & 83.0  & 45.0   &  15.9    & 74.2 & 51.2 & 30.4       & 65.9   \\
PointPainting\cite{vora2020pointpainting} & 46.4 & 58.1  & 77.9  & 35.8   & 15.8  & 73.3 & 41.5 & 24.1   & 60.2   \\
MoCa\cite{moca}          & 66.6 & 70.9  & 86.7 & 58.6    & 32.6  & 87.1 & 67.8 & 52.0   &  72.3   \\
AutoAlign\cite{autoalign} & 65.8 & 70.9  & 85.9 & 55.3 & 29.6  & 86.4  & 71.5 & 51.5   & - \\
FusionPainting\cite{fusionpainting} & 68.1 & 71.6  & 87.1 & 60.8    & 30.0  & 88.3 & 74.7 & 53.5   & 71.8   \\
TransFusion\cite{transfusion} & 68.9 & 71.7  & 87.1 & 60.0  & 33.1   & 88.4 & 73.6 & 52.9   & 78.1 \\
Focals Conv\cite{focalsconv}    & 67.8 & 71.8 & 86.5 & 57.5 & 31.2  & 87.3 & 76.4 & 52.5  & 72.3   \\
VFF\cite{vff} & 68.4 & 72.4 & 86.8 & 58.1 & 32.1 & 87.1 & 78.5 & 52.9 & 73.9 \\
AutoAlignv2\cite{autoalignv2} & 68.4 & 72.4  & 87.0 & 59.0 & 33.1  & 87.6  & 72.9 & 52.1   & 78.0 \\
BEVFusion\cite{bevfusion} & 70.2 & 72.9  & \textbf{88.6} & 60.1 & \textbf{39.3}  & \textbf{89.2}  & 74.1 & 51.0   & \textbf{80.0} \\

\cline{1-10}
\textbf{3D Dual-Fusion (C)}      & 67.9 & 71.5  & 87.5 & 58.8  & 31.6  & 86.9 & 76.6 & 54.9  & 67.3   \\
\textbf{3D Dual-Fusion (T)}      & \textbf{70.6} & \textbf{73.1}  & 87.7 & \textbf{62.0} & 35.6 & 88.8 & \textbf{78.0} & \textbf{58.0}  & 74.4
  \\
\cline{1-10}
\label{table:nus_test}
\end{tabular}
\end{table*}

%% file: table/abl_overall.tex
\begin{table*}[t]
\renewcommand{\arraystretch}{1.0}
\centering
\caption{Ablations for evaluating main modules on KITTI / nuScenes \textit{val} split.}
    \begin{tabular}[t]{c|cccc|ccc|cc}
    \cline{1-10}
     &  &  &  &  & \multicolumn{3}{c|}{KITTI}  &  \multicolumn{2}{c}{nuScenes}   \\
    \cline{1-10}
    \textit{Method} & \textit{naive fusion} & DDA & 3D-LSA & AGFN &  Easy     & Mod   & Hard &  mAP  & NDS  \\
    \cline{1-10}
    \multirow{1}{*}{Baseline\cite{voxelrcnn,centerpoint}} &  &   &    &   & 89.41    & 84.52 & 78.93 & 53.1 & 61.3 \\
    \cline{1-10}
    \multirow{5}{*}{3D Dual-Fusion}&  \multicolumn{1}{c}{\checkmark} &  & &    & 92.16    & 85.14 & 83.14 & 58.2 & 63.2 \\

    & \multicolumn{1}{c}{\checkmark}  & \multicolumn{1}{c}{\checkmark} &     &   & 92.21    & 85.61 & 83.23 & 61.1 & 65.2 \\
    & \multicolumn{1}{c}{\checkmark} & \multicolumn{1}{c}{\checkmark}&\multicolumn{1}{c}{\checkmark}& & 92.46    & 85.90 & 83.47 & 61.5 & 65.9 \\
    & \multicolumn{1}{c}{\checkmark} & \multicolumn{1}{c}{\checkmark}&&\multicolumn{1}{c|}{\checkmark} & 92.26    & 85.66 & 83.21 & 61.8 & 66.0 \\
    & \multicolumn{1}{c}{\checkmark} & \multicolumn{1}{c}{\checkmark}&\multicolumn{1}{c}{\checkmark}& \multicolumn{1}{c|}{\checkmark}& \textbf{92.80}    & \textbf{85.96} & \textbf{83.62} & \textbf{62.1} & \textbf{66.2} \\
    
    \cline{1-10}
    \end{tabular}
    \label{table:abl_overall}

\end{table*}

%% file: table/nus_val.tex
\begin{table*}[htb]

\renewcommand{\arraystretch}{1.0}
\centering
\caption{Performance comparison of 3D object detectors on nuScenes \textit{val} set. }
\begin{tabular}{ccccc}
\cline{1-5}
\textit{Method} &Img Backbone & LiDAR Backbone  & mAP  & NDS \\
\cline{1-5}                                                         
\multicolumn{5}{c}{\textit{LiDAR-based}}  \\
\cline{1-5}                 
SECOND\cite{second}   & - & VoxelNet  & 50.9  & 62.0  \\
CenterPoint\cite{centerpoint}   & - & VoxelNet  & 59.6  & 66.8  \\
FocalsConv \cite{focalsconv} & - & VoxelNet-FocalsConv & 61.2 & 68.1 \\
TransFusion-L\cite{transfusion}   & - & VoxelNet  & 65.1  & 70.1 
\\

\cline{1-5}               
\multicolumn{5}{c}{\textit{LiDAR-Camera based}}  \\
\cline{1-5}            
FUTR3D\cite{futr3d}   & R101 & VoxelNet  & 64.5  & 68.3  \\
FocalsConv \cite{focalsconv} & R50 & VoxelNet-FocalsConv & 65.6 & 70.4 \\
MVP\cite{mvp}   & DLA34 & VoxelNet  & 67.1  & 70.8  \\
AutoAlignv2\cite{autoalignv2}   & CSPNet & VoxelNet  & 67.1  & 71.2  \\
TransFusion\cite{transfusion}   & R50 & VoxelNet  & 67.5  & 71.3  \\
BEVFusion\cite{bevfusion}   & Swin-Tiny & VoxelNet  & 67.9  & 71.0  \\
BEVFusion\cite{bevfusion2}   & Swin-Tiny & VoxelNet  & 68.5  & 71.4  \\
\cline{1-5}
\textbf{3D Dual-Fusion (C)}      & R50 & VoxelNet  & 67.3  & 71.1  \\
\textbf{3D Dual-Fusion (T)}      & R50 & VoxelNet  & \textbf{69.3}  & \textbf{72.2}  \\
\cline{1-5}
\label{table:nus_val}
\end{tabular}
\end{table*}

%% file: table/abl_nus_dist.tex
\begin{table}[t]
\centering
\caption{NDS performance vs. distance on nuScenes $1/7$. "3D DF" indicates 3D Dual-Fusion (C)}
\begin{tabular}{cccc}
\cline{1-4}
\cline{2-4}
\textit{Method} & ${<}$15m  & 15-30m      & ${>}$30m  \\
\cline{1-4}
Baseline\cite{centerpoint}& 71.7 & 62.1 & 38.4 \\
3D DF& 73.8(+2.1) & 67.6(+5.5) & 43.9(+5.5)\\

\cline{1-4}
\label{table:abl_nus_dist}

\end{tabular}
\end{table}

%% file: table/abl_position_encoding.tex
\begin{table}[t]
\renewcommand{\arraystretch}{1.0}
\centering
\caption{Comparison of different positional encoding methods on nuScenes $1/7$. ”DDA” indicates Dual-query Deformable Attention.}
\begin{tabular}{cccc}
\cline{1-4}
\textit{Method}    & PE method      & mAP  & NDS  \\
\cline{1-4}
Baseline\cite{centerpoint}& & 53.1 & 61.3 \\
\cline{1-4}
\multirow{2}{*}{DDA}  
         & Image coordinates    & 60.6 & 64.6 \\
        & Depth        & \textbf{62.1} & \textbf{66.2} \\
\cline{1-4}
\label{table:abl_pe}
\end{tabular}
\vspace{-0.5cm}
\end{table}

%% file: table/abl_num_encoder_layer_kitti.tex
\begin{table}[t]
\renewcommand{\arraystretch}{1.0}
\centering
\caption{Comparison of number of decoder layers on KITTI \textit{val} split.}

    \centering
    \begin{tabular}{cccc}
    \cline{1-4}
    \# of decoder layers & Easy  & Mod   & Right \\ 
    \cline{1-4}
    Baseline\cite{voxelrcnn}       & 89.41 & 84.52 & 78.93 \\
    \cline{1-4}
    1                    & 92.51 & 85.38 & 83.21 \\
    2                    & 92.32 & 85.5  & 83.21 \\
    3                    & 92.52 & 85.66 & 83.29 \\
    \textbf{4}                    & \textbf{92.80} & \textbf{85.96} & \textbf{83.62} \\
    \cline{1-4}
    \end{tabular}
\label{table:abl_num_encoder_layer_kitti}
\vspace{-0.2cm}
\end{table}

%% file: table/abl_num_encoder_layer_nus.tex
\begin{table}[htb!]
\renewcommand{\arraystretch}{1.0}
\centering
\caption{Comparison of number of decoder layers on nuScenes \textit{val} split.}

    \centering
    \begin{tabular}{ccc}
    \cline{1-3}
    \# of decoder layers & mAP  & NDS   \\
    \cline{1-3}
    Baseline\cite{centerpoint}       & 53.1 & 61.3 \\
    \cline{1-3}
    1                    & 60.8 & 65.5 \\
    \textbf{2}                    & \textbf{62.1} & \textbf{66.2} \\
    3                    & 61.9 & 65.8 \\
    \cline{1-3}
    \end{tabular}

\label{table:abl_num_encoder_layer_nus}
\end{table}

%% file: table/abl_query_type_nus.tex
\begin{table}[t]
\centering
\caption{Comparison of different query types on nuScenes $1/7$.}
\begin{tabular}{cccc}
\cline{1-4}
\textit{Method}    & Query type      & mAP  & NDS  \\
\cline{1-4}
Baseline\cite{centerpoint}& & 53.1 & 61.3 \\
\cline{1-4}
\multirow{2}{*}{DDA}  
         & Single camera query    & 59.7 & 65.1 \\
                & Dual query     & \textbf{62.1} & \textbf{66.2}\\

\cline{1-4}
\label{table:abl_query_type}
\end{tabular}
\end{table}

%% file: table/abl_dda_fusion_method.tex
\begin{table}[t]
\centering
\caption{Comparison of different dual-query fusion methods on nuScenes $1/7$. }
\begin{tabular}{cccc}
\cline{1-4}
\textit{Method}    & Fusion method      & mAP  & NDS  \\
\cline{1-4}
Baseline\cite{centerpoint}& & 53.1 & 61.3 \\
\cline{1-4}
\multirow{3}{*}{DDA}  
         & Concat    & 59.5 & 64.1 \\
        &Sum        & 60.2 & 64.0 \\
                &Gated fusion     & \textbf{62.1} & \textbf{66.2}\\

\cline{1-4}
\label{table:abl_dda_fm}
\end{tabular}
\end{table}